\def\year{2020}\relax
\documentclass[letterpaper]{article} % DO NOT CHANGE THIS
\usepackage{aaai20}  % DO NOT CHANGE THIS
\usepackage{times}  % DO NOT CHANGE THIS
\usepackage{helvet} % DO NOT CHANGE THIS
\usepackage{courier}  % DO NOT CHANGE THIS
\usepackage[hyphens]{url}  % DO NOT CHANGE THIS
\usepackage{graphicx} % DO NOT CHANGE THIS
\usepackage{graphicx}  % DO NOT CHANGE THIS
\usepackage{tabularx} % extra features for tabular environment
\usepackage{array}
\newcolumntype{L}[1]{>{\raggedright\let\newline\\\arraybackslash\hspace{0pt}}m{#1}}
\newcolumntype{C}[1]{>{\centering\let\newline\\\arraybackslash\hspace{0pt}}m{#1}}
\newcolumntype{R}[1]{>{\raggedleft\let\newline\\\arraybackslash\hspace{0pt}}m{#1}}

\usepackage[hang,flushmargin]{footmisc} % footnote indent

\usepackage{amsmath}  % improve math presentation
\usepackage{graphicx} % takes care of graphic including machinery
\usepackage{mathtools}
\usepackage{amssymb}
\usepackage{booktabs} % For formal tables
\usepackage{multirow}
\usepackage{adjustbox}
\usepackage{makecell}
\usepackage{threeparttable}
\usepackage{subfigure}
\usepackage{overpic}
\usepackage{bm} % bm
\usepackage{soul}

\usepackage{balance} % balance references
\usepackage{booktabs} % professional table
\usepackage{threeparttable} % table footnote
\usepackage{enumitem} % enumerate

\graphicspath{{Figs/}}

\newcommand{\norm}[1]{\left\lVert#1\right\rVert}
\DeclareMathOperator*{\argmin}{argmin}

\newcommand{\RNum}[1]{\uppercase\expandafter{\romannumeral #1\relax}}

\newsavebox\CBox

\title{MALA: Cross-Domain Dialogue Generation with Action Learning}
 \author{Xinting Huang,\textsuperscript{1}
Jianzhong Qi,\textsuperscript{1}
Yu Sun,\textsuperscript{2}
Rui Zhang\textsuperscript{1}\thanks{Rui Zhang is the corresponding author.}\\
\textsuperscript{1}{The University of Melbourne}, 
\textsuperscript{2}{Twitter Inc.}\\
\{xintingh@student., jianzhong.qi@, rui.zhang@\}unimelb.edu.au,
ysun@twitter.com}

\begin{document}

\maketitle

\begin{abstract}
Response generation for task-oriented dialogues involves two basic components: dialogue planning and surface realization. 
These two components, however, have a discrepancy in their objectives, i.e., task completion and language quality. 
To deal with such discrepancy, conditioned response generation has been introduced where the generation process is factorized into action decision and language generation via explicit action representations. 
To obtain action representations, recent studies learn \emph{latent actions} in an unsupervised manner based on the utterance lexical similarity.
Such an action learning approach is prone to diversities of language surfaces, which may impinge task completion and language quality.
To address this issue, we propose \emph{multi-stage adaptive latent action} learning (MALA) that learns \emph{semantic latent actions} by distinguishing the effects of utterances on dialogue progress.
We model the utterance effect using the transition of dialogue states caused by the utterance and develop a semantic similarity measurement that estimates whether utterances have similar effects.  
For learning semantic actions on domains without dialogue states, MALA extends the semantic similarity measurement across domains progressively, i.e., from aligning shared actions to learning domain-specific actions.
Experiments using multi-domain datasets, SMD and MultiWOZ, show that our proposed model achieves consistent improvements over the baselines models in terms of both task completion and language quality.
% To model the utterance effect, we utilize the transition of dialogue states after this utterance, since dialogue states summarize the current progress of dialogues.

\end{abstract}

\section{Introduction}

Task-oriented dialogue systems complete tasks for users, such as making a restaurant reservation or scheduling a meeting, in a multi-turn conversation \cite{gao2018neural,sun2016contextual,sun2017collaborative}.
Recently, end-to-end approaches based on neural encoder-decoder structure have shown promising results \cite{wen2017network,madotto2018mem2seq}.
However, such approaches directly map plain text dialogue context to responses (i.e., utterances), and do not distinguish two basic components for response generation: \emph{dialogue planning} and \emph{surface realization}.
% \emph{what to say} and \emph{how to say} for response generation.
Here, dialogue planning means choosing an action (e.g., to request information such as the preferred cuisine from the user, or provide a restaurant recommendation to the user), and surface realization means transforming the chosen action into natural language responses.
Studies show that not distinguishing these two components can be problematic since they have a discrepancy in objectives, and optimizing decision making on choosing actions might adversely affect the generated language quality \cite{yarats2018hierarchical,zhao2019rethinking}.

\begin{table} [!tbp]
    \centering
    
\begin{threeparttable}
    \centering
\caption{ \small System Utterance Action Example 
} \label{toy-example}
% \tiny
%\scriptsize
%\footnotesize
\small
% \normalsize  
\setlength\tabcolsep{3.0pt}

\begin{tabular}{L{4.0cm}|L{4.0cm}}
\toprule
\multicolumn{2}{c}{\textbf{System utterances}} \\
\midrule
Domain: Hotel \newline
(a). \textit{Was there a particular section of town you were looking for?} \newline (b). \textit{Which area could you like the hotel to be located at?} & Domain: Attaction \newline (c). \textit{Did you have a particular type of attraction you were looking for?} \newline (d). \textit{great , what are you interested in doing or seeing ?}\\
\midrule
\multicolumn{2}{c}{\textbf{System intention (ground truth action)}} \\
\midrule
\texttt{Request(Area)} &{\texttt{Request(Type)}}\\
\midrule
\multicolumn{2}{c}{\textbf{Latent action (auto-encoding approach) }} \\
\midrule
(a): \textit{[0,0,0,1,0]}; (b): \textit{[0,1,0,0,0]} & (c): \textit{[0,0,0,1,0]}; (d): \textit{[0,0,0,0,1]} \\
\midrule
\multicolumn{2}{c}{\textbf{Semantic latent action (proposed)}}  \\
\midrule
(a) \& (b): \textit{[0,0,0,1,0]} & (c) \& (d): \textit{[0,0,0,0,1]} \\
\bottomrule
\end{tabular}

% \begin{tablenotes}\footnotesize
% \item[*]
% \end{tablenotes}
\end{threeparttable}

\end{table}

% Are you interested in having dinner? going to a bar?

To address this problem, conditioned response generation that relies on action representations has been introduced \cite{wen2015semantically,chen2019semantically}.
Specifically, each system utterance is coupled with an explicit action representation, and responses with the same action representation convey similar meaning and represent the same action.
In this way, the response generation is decoupled into two consecutive steps, and each component for conditioned response generation (i.e., dialogue planning or surface realization) can optimize for different objectives without impinging the other.
Obtaining action representations is critical to conditioned response generation.
Recent studies adopt variational autoencoder (VAE) to obtain low-dimensional latent variables that represent system utterances in an unsupervised way.
% the diversities of languages surfaces
Such an auto-encoding approach cannot effectively handle various types of surface realizations, especially when these exist multiple domains (e.g., hotel and attraction).
This is because the latent variables learned in this way mainly rely on the lexical similarity among utterances instead of capturing the underlying intentions of those utterances. 
In \mbox{Table \ref{toy-example}}, for example, system utterances (a) and (c) convey different intentions (i.e., \texttt{request(area)} and \texttt{request(type)}), but may have the same auto-encoding based latent action representation since they share similar wording.

To address the above issues, we propose a multi-stage approach to learn \emph{semantic latent actions} that encode the underlying intention of system utterances instead of surface realization.
The main idea is that the system utterances with the same underlying intention (e.g., \texttt{request(area)}) will lead to similar \emph{dialogue state transitions}.
% explain dialogue state and state transition 
This is because dialogue states summarize the dialogue progress towards task completion, and a dialogue state transition reflect how the intention of system utterance influences the progress at this turn. 
% the user feedbacks (e.g., inform more requirements) to the system utterance.
% an example? (probably not)
To encode underlying intention into semantic latent actions, we formulate a loss based on whether the reconstructed utterances from VAE cause similar state transitions as the input utterances.
% To capture underlying intention more effectively
To distinguish the underlying intention among utterances more effectively, we further develop a regularization based on the similarity of resulting state transitions between two system utterances.

% A requirement of obtaining the semantic latent actions is annotations of the dialogue states.
Learning the semantic latent actions requires annotations of the dialogue states.
In many domains, there are simply no such annotations because they require extensive human efforts and are expensive to obtain.
% referred to as target domains,
We tackle this challenge by transferring the knowledge of learned semantic latent actions from state annotation rich domains (i.e., source domains) to those without state annotation (i.e., target domains).
% using a \emph{easy-to-hard} strategy
We achieve knowledge transferring in a progressive way, and start with actions that exist on both the source and target domain, e.g., \texttt{Request(Price)} in both hotel and attraction domain.
We call such actions as \emph{shared actions} and actions only exist in the target domain as \emph{domain-specific actions}.  
We observe that system utterances with shared actions will lead to similar states transitions despite belonging to different domains. 
% and the resulting dialogue states transitions are similar despite belonging to different domains.
% when placing utterances with the same shared action in the same dialogue context 
Following this observation, we find and align the shared actions across domains.
% (i.e., semantic latent action)
With action-utterance pairs gathered from the above shared actions aligning, we train a network to predict the similarity of resulting dialogue state transitions by taking as input only texts of system utterances. 
We then use such similarity prediction as supervision to better learn semantic latent actions for all utterances with domain-specific actions.

% [topsep=0pt,leftmargin=*,noitemsep,wide=0pt]
Our contributions are summarized as follows: 
\begin{itemize}[topsep=0pt,leftmargin=*,noitemsep,wide=0pt]
\item 
We are the first to address the problem of cross-domain conditioned response generation without requiring action annotation. 
\item 
We propose a novel latent action learning approach for conditioned response generation which captures underlying intentions of system utterances beyond surface realization.
\item
We propose a novel multi-stage technique to extend the latent action learning to cross-domain scenarios via shared-action aligning and domain-specific action learning.
\item 
We conduct extensive experiments on two multi-domain human-to-human conversational datasets. The results show the proposed model outperforms the state-of-the-art on both in-domain and cross-domain response generation settings.
\end{itemize}

\section{Related Work}

\subsection{Controlled Text Generation}
Controlled text generation aims to generate responses with controllable attributes. 
Many studies focus on open-domain dialogues' controllable attributes, e.g., style \cite{yang2018unsupervised}, sentiment \cite{shen2017style}, and specificity \cite{zhang2018learning}.
% diversity \cite{zhao2017learning}
Different from open-domain, the controllable attributes for task-oriented dialogues are usually \emph{system actions}, since it is important that system utterances convey clear intentions.
Based on handcrafted system actions obtained from domain ontology, action-utterance pairs are used to learn semantically conditioned language generation models \cite{wen2015semantically,chen2019semantically}. 
Since it requires extensive efforts to build action sets and collect action labels for system utterances, recent years have seen a growing interest in learning utterance representations in an unsupervised way, i.e., \emph{latent action learning} \cite{zhao2018unsupervised,zhao2019rethinking}.
Latent action learning adopts a pretraining phase to represent each utterance as a latent variable using a reconstruction based variational auto-encoder \cite{yarats2018hierarchical}.
The obtained latent variable, however, mostly reflects lexical similarity and lacks sufficient semantics about the intention of system utterances. 
% Besides reconstruction based action learning,
We utilize the dialogue state information to enhance the semantics of the learned latent actions.  

% Besides the lexical similarity, we also utilize the dialogue state information to enhance the state semantics of the learned latent actions.

% Specifically, we utilize dialogue state information to measure utterance effects on dialogue progress, 
% thus utterances with the same action convey the same meaning instead of lexical similarity.
% and incorporate such effect modelling into action learning.

\subsection{Domain Adaptation for Task-oriented Dialogues}

Domain adaptation aims to adapt a trained model to a new domain with a small amount of new data.
This is studied in computer vision \cite{saito2017asymmetric}, item ranking \cite{wang2018joint,huang2019carl}, and multi-label classification \cite{wang2018kdgan,wang2019adversarial,sun2019internet}. 
% This is essential for task-oriented dialogues due to data scarcity.
For task-oriented dialogues, early studies focus on domain adaptation for individual components,
e.g., intention determination \cite{chen2016zero}, dialogue state tracking \cite{mrkvsic2015multi}, and dialogue policy \cite{mo2018personalizing,yin2018context}.
Two recent studies investigate end-to-end domain adaptation.
DAML \cite{qian2019domain} adopts model-agnostic meta-learning to learn a seq-to-seq dialogue model in target domains. 
ZSDG \cite{zhao2018zero} conducts adaptation based on action matching, and uses partial target domain system utterances as domain descriptions.
These end-to-end domain adaption methods are either difficult to be adopted for conditioned generation or needing a full annotation of system actions.
We aim to address these limitations in this study.

\begin{figure}[!t]
\centering
\subfigure[\footnotesize{Stage-I: Semantic Latent Action Learning}]{
\begin{overpic}[height=2.94cm]{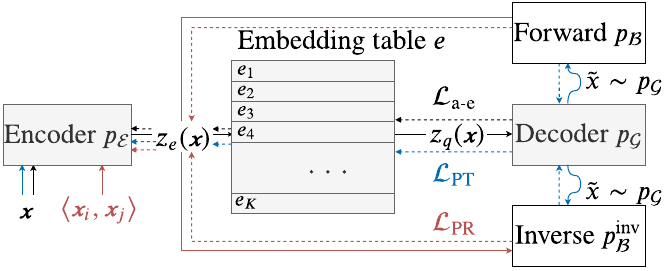}
\label{subfig-stage1}
\end{overpic}
}
\subfigure[\footnotesize{Stage-II: Action Alignment across Domains}]{
\begin{overpic}[height=2.94cm]{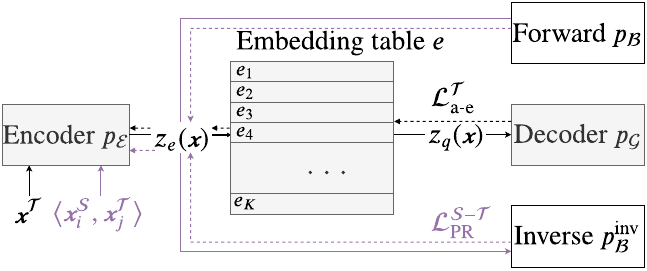}
\label{subfig-stage2}
\end{overpic}
}
\subfigure[\footnotesize{Stage-III: Domain-Specific Action Learning}]{
\begin{overpic}[height=2.54cm]{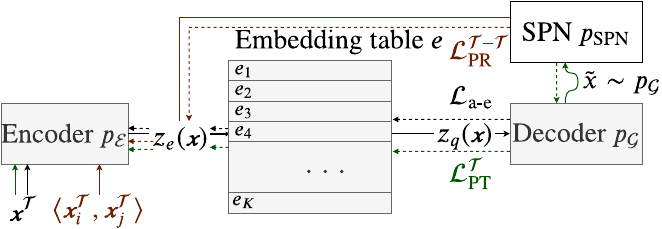}
\label{subfig-stage3}
\end{overpic}
}
\caption{Overall Framework of MALA.}
\label{fig-overall}
\end{figure}

\section{Preliminaries}

Let $\{d_i | 1 \leq i \leq N \} $ be a set of dialogue data, and each dialogue $d_i$ contains $n_d$ turns: $d_i = \{ (c_t, x_t) | 1\leq t \leq n_d \}$, 
where $c_t$ and $x_t$ are the context and system utterance at turn $t$, respectively. 
The context $c_t =\{u_1, x_1,...u_t \}$ consists of the dialogue history of user utterances $u$ and system utterances $x$.
Latent action learning aims to map each system utterance $x$ to a representation $z_d(x)$, where utterances with the same representation express the same action.  
The form of the representations $z_d(x)$ can be, e.g., one-hot \cite{wen2015semantically}, multi-way categorical, and continuous \cite{zhao2019rethinking}.
We use the one-hot representation due to its simplicity although the proposed approach can easily extend to other representation forms. 
% , i.e., $z_d(x)\in [K]$, where $[K] = \{1,2, ..., K\}$,

% a vector quantization based discretization bottleneck
We obtain the one-hot representation via VQ-VAE, a discrete latent VAE model \cite{van2017neural}.
Specifically, an encoder $p_{\mathcal{E}}$ encodes utterances as $z_e(x) \in \mathbb{R}^D$, and a decoder $p_{\mathcal{G}}$ reconstructs the original utterance based on inputs $z_q(x) \in \mathbb{R}^D$, where $D$ is the hidden dimension. 
The difference lies in that between $z_e(x)$ and $z_q(x)$, we build a discretization bottleneck using a nearest-neighbor lookup on an embedding table $e \in \mathbb{R}^{K \times D}$ and obtain $z_q(x)$ by finding the embedding vector in $e$ having the closest Euclidean distance to $z_e(x)$ i.e., 
\begin{equation*}
 z_q(x) =e_k \text{   where   }    k=\argmin_{j \in \mid K \mid} \norm{z_e(x)-e_j}_2.
\end{equation*}
The learned latent $z_d(x)$ is a one-hot vector that only has 1 at index $k$.    
All components, including $p_{\mathcal{E}}$, $p_{\mathcal{G}}$ and embedding table $e$, are jointly trained using \underline{a}uto-\underline{e}ncoding objective as
% \begin{equation}
% \begin{split}
%     \mathcal{L}_{\text{a-e}} =\mathbb{E}_{ \vect{x}} & [-\log p_{\mathcal{G}}(x|z_q(x)) +\norm{\text{sg}(z_e(\vect{x}) )-z_q(\vect{x})}_2^2 \\
%     &+ \norm{z_e(x) -\text{sg}(e) }_2^2]
% \end{split}
% \end{equation}
\begin{equation}
    \mathcal{L}_{\text{a-e}} =\mathbb{E}_{ x} [-\log p_{\mathcal{G}}( x|z_q(  x)) + \norm{z_e(x )-z_q(x)}_2^2]
\end{equation}
% where $\text{sg}(\cdot)$ is the stop gradient operator.
The structure of VQ-VAE is illustrated in Fig. \ref{subfig-stage1}, where the three components are marked in grey color.  
% and how $\mathcal{L}_{\text{a-e}}$ is used for training

\section{Proposed Model}

\subsection{Overview}
To achieve better conditioned response generation for task-oriented dialogues, we propose \emph{\underline{m}ulti-stage \underline{a}daptive \underline{l}atent \underline{a}ction learning} (MALA).
Our proposed model works for two scenarios: 
(i) For domains with dialogue state annotations, we utilize these annotations to learn semantic latent actions to enhance the conditioned response generation.
% by modeling dialogue state transition influenced by system intentions.
% by encoding into 
(ii) For domains without state annotations, we transfer the knowledge of semantic latent actions learned from the domains with rich annotations, and thus can also enhance the conditioned response generation for these domains.

The overall framework of MALA is illustrated in \mbox{Fig. \ref{fig-overall}}.
The proposed model is built on VQ-VAE that contains encoder $p_{\mathcal{E}}$, embedding table $e$, and decoder $p_{\mathcal{G}}$. 
Besides auto-encoding based objective $\mathcal{L}_{\text{a-e}}$, we design \ul{p}oin\ul{t}wise loss $\mathcal{L}_{\text{PT}}$ and \ul{p}ai\ul{r}wise loss $\mathcal{L}_{\text{PR}}$ to enforce the latent actions  to reflect underlying intentions of system utterances. 
% Based on whether state annotations are available, we choose among state tracking models ($p_{\mathcal{B}}$ and $p_{\mathcal{B}}^{\text{inv}} $) or similarity prediction network ($p_{\text{SPN}}$) to provide supervision.
For domains with state annotations (see Fig. 1a), we train $p_{\mathcal{B}}$ and $p_{\mathcal{B}}^{\text{inv}} $ to measure state transitions and develop the pointwise and pairwise loss (Sec. 4.2).
For domains without state annotations (see Fig. 1b), we develop a pairwise loss $\mathcal{L}_{\text{PR}}^{\mathcal{S-T}} $ based on $p_{\mathcal{B}}$ and $p_{\mathcal{B}}^{\text{inv}} $ from annotation-rich-domains.
This loss measure state transitions for a cross-domain utterance pair, and thus can find and align shared actions across domains (Sec. 4.3).
%whether two utterance from different domains can lead to similar state transitions, 
We then train a similarity prediction network $p_{\text{SPN}}$ to substitute the role of state tracking models, which only taking as input raw text of utterances.
We using $p_{\text{SPN}}$ predictions as supervision to form pointwise $\mathcal{L}_{\text{PT}}^{\mathcal{T-T}}  $ and pairwise loss $\mathcal{L}_{\text{PR}}^{\mathcal{T-T}}  $ (see Fig. 1c), and thus obtain semantic latent actions for domain without state annotations (Sec. 4.4).

\iffalse
The overall framework of MALA is illustrated in \mbox{Fig. \ref{fig-overall}}.
We first utilize dialogue states to obtain semantic latent actions for the state-annotation-rich domains (Sec. \ref{stage1}). 
For domains having no state annotations, we adopt an progressive strategy to transfer the knowledge of semantic latent actions.
We first find and align \emph{shared actions} across domains (Sec. \ref{stage2}).
% After that, based on domain knowledge gathered on action alignment stage, we train a similarity prediction network (SPN) and 
We then further learn fine-grained \emph{domain-specific actions} by extending the shared semantic latent actions to the complete action space (Sec. \ref{stage3}).
\fi

\iffalse
The proposed model is built on VQ-VAE that contains encoder $p_{\mathcal{E}}$, embedding table $e$, and decoder $p_{\mathcal{G}}$. Besides auto-encoding based objective $\mathcal{L}_{\text{a-e}}$, we design pointwise loss $\mathcal{L}_{\text{PT}}$ and pairwise $\mathcal{L}_{\text{PR}}$ to enforce the learned discrete action representations to reflect underlying intentions of system utterances. Based on whether state annotations are available, we choose among state tracking models ($p_{\mathcal{B}}$ and $p_{\mathcal{B}}^{\text{inv}} $) or similarity prediction network ($p_{\text{SPN}}$) to provide supervision.
\fi

\subsection{Stage-I: Semantic Latent Action Learning}\label{stage1}
We aim to learn semantic latent actions that align with the underlying intentions for system utterances.
To effectively capture the underlying intention, we utilize dialogue state annotations and regard utterances that lead to similar state transition as having the same intention.
We train dialogue state tracking model to measure whether any two utterance will lead to a similar state transition. 
We apply such measurement in (i) a pointwise manner, i.e., between a system utterance and its reconstructed counterpart from VAE, and (ii) a pairwise manner, i.e., between two system utterances.   
% to model system utterance's effects and further develop a semantic similarity measurement.

% Thus, we can obtain semantic latent actions by incorporating such similarity measurement into a self-supervised discrete representation learning framework.
% We detail this latent action learning stage in this section.

\subsubsection{Dialogue State Tracking} 
Before presenting the proposed pointwise measure, we first briefly introduce dialogue state tracking tasks. 
Dialogue states (also known as dialogue belief) are in the form of predefined slot-value pairs. 
Dialogues with state (i.e., belief) annotations are represented as $d_i = \{ (c_t, b_t, x_t) | 1\leq t \leq n_d \} $, where $b_t \in \{0,1\}^{N_b}$ is the dialogue state at turn $t$, and $N_b$ is the number of all slot-value pairs.
Dialogue state tracking (DST) is a multi-label learning process that models the conditional distribution $p(b_t|c_t)=p(b_t|u_t, x_{t-1}, c_{t-1})$.    
Using dialogue state annotations, we first train a state tracking model $p_{\mathcal{B}}$ with the following cross-entropy loss:
\begin{equation}
\begin{split}
    \mathcal{L} = \sum_{d_i}\sum_{t=1:n_d}-\log(b_t^{\top} \cdot p_{\mathcal{B}}(u_t, x_{t-1}, c_{t-1})) \\
\end{split}
\end{equation}
\begin{equation*}
     p_{\mathcal{B}}(u_t, x_{t-1}, c_{t-1}) =  \text{softmax}(h(u_t, x_{t-1}, c_{t-1}))
\end{equation*}
where $h(\cdot)$ is a scoring function and can be implemented in various ways, e.g., a self attention model \cite{zhong2018global}, or an encoder-decoder \cite{wu2019transferable}.

\iffalse
For convenience, we refer to domains with dialogue state information as \emph{source domains} and domains without state information as \emph{target domains}. 
Note that the DST model is only available in the source domain because target domain has no labeled dialogue states. 
\fi

\subsubsection{Pointwise Measure}
\iffalse
Built on unsupervised latent action learning framework (i.e., VAE with discretization bottleneck), we overcome its limitation in capturing underlying intention by additionally encoding pointwise and pairwise measures beyond lexical similarity.
\fi

% In order to overcome limitations of auto-encoding based latent action learning, 
% considering surface realization only, 
% we consider utterances effects on dialogue progresses for latent action learning.
% To achieve this, we develop a pointwise effect measure using dialogue state information.

With the trained state tracking model $p_{\mathcal{B}}$, we now measure whether the reconstructed utterance output can lead to a similar dialogue state transition from turn $t-1$ to $t$ (i.e., forward order).
We formulate such measure as a cross-entropy loss between original state $b_t$ and model $p_{\mathcal{B}}$ outputs when replacing system utterance $x_{t-1}$ in inputs with $\Tilde{x}_{t-1}$
\begin{equation}
    \mathcal{L}_{\text{fwd}} = \mathbb{E}_{x}[-\log (b_t^{\top} \cdot p_{\mathcal{B}}(b_t|u_t, \Tilde{x}_{t-1}, c_{t-1}) )]
\end{equation}
\begin{equation*}
    \Tilde{x}_{t-1} \sim p_{\mathcal{G}} (z_q(x_{t-1}))
\end{equation*}
where $\Tilde{x}_{t-1}$ is sampled from the decoder output. 
Note that once state tracking model $p_{\mathcal{B}}$ finish training, its parameters will not be updated and $\mathcal{L}_{\text{fwd}}$ is only used for training the components of VAE, i.e., the encoder, decoder and the embedding table.
To get gradients for these components during back-propagation, we apply a continuous approximation trick \cite{yang2018unsupervised}. 
% add technical details of the approximation trick.
Specifically, instead of feeding sampled utterances as input to state tracking models, we use Gumbel-softmax \cite{jang2016categorical} distribution to sample instead. 
In this way outputs of the decoder $p_{\mathcal{G}} $ becomes a sequence of probability vectors, and we can use standard back-propagation to train the generator.

We expect the dialogue state transition in forward order can reflect the underlying intentions of system utterances.
However, the state tracking model $p_{\mathcal{B}}$ heavily depends on user utterance $u_t$, meaning that shifts of system utterance intentions may not sufficiently influence the model outputs. 
This prevents the considered state transitions modeled from providing valid supervision for semantic latent action learning.
To address this issue, inspired by inverse models in reinforcement learning \cite{pathak2017curiosity}, we formulate inverse state tracking to model the dialogue state transition from turn $t$ to $t-1$.
Since dialogue state at turn $t$ already encodes information of user utterance $u_t$, we formulate the inverse state tracking as $p(b_{t-1}|x_{t-1}, b_{t})$.
In this way the system utterance plays a more important role in determining state transition. 
Specifically, we use state annotations to train an inverse state tracking model $p_{\mathcal{B}}^{\text{inv}}$ using the following cross-entropy loss:
\begin{equation}
    \mathcal{L} = \sum_{d_i}\sum_{t=2:n_d}-\log(b_{t-1}^{\top} \cdot p_{\mathcal{B}}^{\text{inv}}(|x_{t-1}, b_t))
\end{equation}
\begin{equation*}
    p_{\mathcal{B}}^{\text{inv}}(x_{t-1}, b_t) = \text{softmax}(g(x_{t-1}, b_{t-1}))
\end{equation*}
where the scoring function $g(\cdot)$ can be implemented in the same structure as $h(\cdot)$.
The parameters of inverse state tracking model $p_{\mathcal{B}}^{\text{inv}}$ also remain fixed once training is finished. 

We use the inverse state tracking model to measure the similarity of dialogue state transitions caused by system utterance and its reconstructed counterpart. The formulation is similar to forward order:
\begin{equation}
    \mathcal{L}_{\text{inv}} = \mathbb{E}_{x}[-\log(b_{t-1}^{\top} \cdot p_{\mathcal{B}}^{\text{inv}}(b_{t-1}| \Tilde{x} _{t-1}, b_t))]
\end{equation}
\begin{equation*}
    \Tilde{x}_{t-1} \sim p_{\mathcal{G}} (z_q(x_{t-1})).
\end{equation*}

Thus, combining the dialogue state transitions modeled in both forward and inverse order, we get the full pointwise loss for learning semantic latent actions:
\begin{equation}
    \mathcal{L}_{\text{PT}} = \mathcal{L}_{\text{fwd}} + \mathcal{L}_{\text{inv}}
\end{equation}

\subsubsection{Pairwise Measure}
To learn semantic latent actions that can distinguish utterances with different intentions, we further develop a pairwise measure that estimates whether two utterances lead to similar dialogue state transitions.

With a slight abuse of notation, we use $x_{i}$ and $x_{j}$ to denote two system utterances. 
We use $u_i$, $c_i$, $b_i$ to denote the input user utterance, dialogue context, and dialogue state for dialogue state tracking models $p_{\mathcal{B}}$ and $p_{\mathcal{B}}^{\text{inv}}$, respectively.
We formulate a pairwise measurement of state transitions as
% We not only want the distances between utterances with same-action are close, but those with different actions are far away. 
% To this aim, we further develop a pairwise similarity measure based on the pointwise effect measure.
% The pairwise similarity is measured as the amount of change in DST results when replacing $x_i$ with $x_j$.
% putting $x_{j}$ into context of $x_{i}$ :
\begin{equation}
    s_{i,j} = s_{\text{fwd}}(x_{i}, x_{j}) + s_{\text{inv}}(x_{i}, x_{j})
\end{equation}
\begin{equation*}
\begin{split}
    s_{\text{fwd}}(x_{i}, x_{j}) &= \text{KL}(p_{\mathcal{B}}^{\text{fwd}} (u_i, x_i,c_i) \mid \mid p_{\mathcal{B}}^{\text{fwd}}(u_i,x_j,c_i) )  \\
    s_{\text{inv}}(x_{i}, x_{j}) &= \text{KL}(p_{\mathcal{B}}^{\text{inv}}(x_i, b_i) \mid \mid p_{\mathcal{B}}^{\text{inv}}(x_j, b_i) )
\end{split}
\label{sim-fwd-inv}
\end{equation*}
where KL is the Kullback-Leibler divergence. 
Both $p_{\mathcal{B}}$ and $p_{\mathcal{B}}^{\text{inv}}$ take inputs related to $x_i$. 
% e.g., user utterance, dialogue state, except the input system utterance.   
We can understand $s_{i,j} $ in the way that it measures how similar the state tracking results are when replacing $x_i$ with $x_j$ as input to $p_{\mathcal{B}}$ and $p_{\mathcal{B}}^{\text{inv}} $.  
% , thus such similarity measurement is \emph{asymmetrical}.

To encode the pairwise measure into semantic latent action learning, we first organize all system utterances in a pairwise way
$ \mathcal{P} =\{\big \langle (x_i, x_j), s_{i,j} \big \rangle  |1 \leq i,j \leq N_u^{\mathcal{S}} \} $
where $N_u^{\mathcal{S}}$ is the total number of system utterances in the domains with state annotations.
We then develop a pairwise loss to incorporate such measure on top of the VAE learning
\begin{equation}\label{pair-s1}
    \mathcal{L}_{\text{PR}} = \sum_{\mathcal{P}} - s_{ij}^{\text{avg}}\log d(x_i, x_j) - (1-s_{ij}^{\text{avg}}) \log (1- d(x_i, x_j))
\end{equation}
\begin{equation*}
    %  s_{ij}^{\text{avg}} &= (s_{i,j} + s_{j,i})/2 \\
    d(x_i, x_j) = \sigma ( -z_e(x_{i})^{\top} z_e(x_{j}) )
\end{equation*}
where $\sigma$ is the sigmoid function, $s_{ij}^{\text{avg}}$ is the average of $s_{i,j}$ and $s_{j,i}$, and $z_e(x) \in \mathbb{R^{D}}$ is encoder $p_{\mathcal{E}}$ outputs. The pairwise loss $\mathcal{L}_{\text{PR}}$ trains $p_{\mathcal{E}}$ by enforcing its outputs of two system utterances to have far distances when these two utterance lead to different state transitions, and vice versa.

The overall objective function of the semantic action learning stage is:
\begin{equation}
    \mathcal{L}_{\text{S-\RNum{1}}} = \mathcal{L}_{\text{a-e}} + \alpha \mathcal{L}_{\text{PT}} + \beta \mathcal{L}_{\text{PR}}
\end{equation}
where $\alpha$ and $\beta $ are hyper-parameters. We adopt $\mathcal{L}_{\text{S-\RNum{1}}}$ to train VAE with discretization bottleneck and obtain utterance-action pair (e.g., utterance (c) and its semantic latent action in Table \ref{toy-example}) that encodes the underlying intentions for each system utterance in the domains with state annotations.

% , and the action here is semantic latent action that encodes the underlying intentions of system utterances. 

\subsection{Stage-II: Action Alignment across Domains}\label{stage2}
In order to obtain utterance-action pairs in domains having no state annotations, we propose to progressively transfer the knowledge of semantic latent actions from those domains with rich state annotations.
At this stage, we first learn semantic latent actions for the utterances that have co-existing intentions (i.e., shared actions) across domains.
% To this aim, we first find and align shared actions at this stage.
% Specifically, we find which utterances in the target domain having co-existing underlying intentions (i.e., shared actions), 
% and build utterance-action pairs for these utterances with semantic latent action learned at \mbox{stage-I.} 

We use $x^{\mathcal{S}} $ and $x^{\mathcal{T}} $ to denote system utterances in the source and target domain, respectively.
The set of all utterances is denoted by:
\begin{equation*}
    U^{\mathcal{S}}=\{x_i^{\mathcal{S}} | 1 \leq i \leq N_u^{\mathcal{S}} \} ; U^{\mathcal{T}} =\{x_j^{\mathcal{T}} |1 \leq j \leq N_u^{\mathcal{T}} \}
\end{equation*}
where $N_u^{\mathcal{S}} $ and $N_u^{\mathcal{T}} $ are the total utterance number in each domain, respectively.
% We assume to have $ N_u^{\mathcal{S}} \gg N_u^{\mathcal{T}} $ since the domains having no state annotations are also usually with a limited number of dialogues.
% We extend the pairwise similarity measure into cross-domain setting to find target domain utterances with shared actions.
We adopt the proposed pairwise measure to find the target domain system utterances that have shared actions with the source domain. 
Based on the assumption that although from different domains, utterances with the same underlying intention are expected to lead to similar state transitions, we formulate the pairwise measure of cross-domain utterance pairs as:
\begin{equation}
    s_{i,j}^{c} = s_{\text{fwd}}(x_{i}^{\mathcal{S}}, x_{j}^{\mathcal{T}} ) + s_{\text{inv}}(x_{i}^{\mathcal{S}}, x_{j}^{\mathcal{T}})
\end{equation}
% in the same way in Eqn. \ref{sim-fwd-inv}.
where $s_{\text{fwd}} $ and $s_{\text{inv}} $ are computed using the trained $p_{\mathcal{B}} $ and $p_{\mathrel{B}}^{\text{inv}}$.
Since it only requires the trained dialogue state tracking models and state annotations related to $x_{i}^{\mathcal{S}}$, this pairwise measure is asymmetrical.
Taking advantage of the asymmetry, this cross-domain pairwise measure can still work when we only have raw texts of dialogues in the target domain.  
% only requires dialogue state information and the trained DST models on the source domain.

We then utilize the cross-domain pairwise for action alignment during latent action learning in the target domain.
We formulate a loss incorporating action alignment
\begin{equation}\label{pair-s2}
\begin{split}
    \mathcal{L}_{\text{PR}}^{\mathcal{S-T}} &= \sum_{x^{S}, x^{T}} - s_{i,j}^{c}\log d(x_i^{\mathcal{S}}, x_j^{\mathcal{T} })\\
    & -(1-s_{i,j}^{c})\log (1-d(x_i^{\mathcal{S} },x_j^{\mathcal{T}}))
\end{split}
\end{equation}
\begin{equation*}
      d(x_i^{\mathcal{S}}, x_j^{\mathcal{T}}) = \sigma( -z_e(x_{i}^{\mathcal{S}})^{\top} z_e(x_{j}^{\mathcal{T}} ) )
\end{equation*}
where $d(x_i^{\mathcal{S}} , x_j^{\mathcal{T} })$ is computed based on outputs of the same encoder $p_{\mathcal{E}}$ from VAE at stage-I.
We also use utterances in the target domain to formulate an auto-encoding loss:
\begin{equation}
    \mathcal{L}_{\text{a-e}}^{\mathcal{T}} = \mathbb{E}_{x\in U^{\mathcal{T}}}[l_r+\norm{\text{sg}(z_e(x) )-z_q(x)}_2].
\end{equation}
The overall objective for the stage-$\text{\RNum{2}}$ is:
\begin{equation}
    \mathcal{L}_{\text{S-\RNum{2}}} = \mathcal{L}_{\text{a-e}}^{\mathcal{T}} + \beta \mathcal{L}_{\text{PR}}^{\mathcal{S-T}}
\end{equation}
where $\beta$ is the hyper-parameter as the same in $\mathcal{L}_{\text{S-I}}$. 
With the VAE trained using $\mathcal{L}_{\text{S-\RNum{2}}} $, 
we can obtain utterance-action pairs for system utterances in the domain having no state annotations.
However, for utterances having domain-specific intentions, their semantic latent actions are still unclear, which is tackled in Stage 3. 
% the obtained latent actions are merely based on surface realization (e.g., utterance (b) may have the auto-encoding based latent action instead of the semantic one after stage-II, if \texttt{Request(Area)} does not appear in other domains). 

% among all obtained pairs, only those coupled with utterances that have shared actions appropriately capture the underlying intentions.

% only utterances having co-existing intentions are coupled with semantic latent actions, while those utterances having domain-specific actions are still coupled with merely auto-encoding based latent actions.    

\subsection{Stage-III: Domain-specific Actions Learning}\label{stage3}
% Based on the collected utterance-action pairs on source and target domains,
We aim to learn semantic latent action for utterances with domain-specific actions at this stage. 

\subsubsection{Similarity Prediction Network (SPN)}
We train an utterance-level prediction model, SPN, to predict whether two utterances lead to similar state transitions by taking as input the raw texts of system utterances only. 
% Since there is no dialogue state available on the target domain and pairwise similarity measure cannot perform well for utterances with domain-specific actions
Specifically, SPN gives a similarity score in $[0,1]$ to an utterance pair:
\begin{equation}
    p_{\text{SPN}}(x_i,x_j) = \sigma(r(x_i, x_j))
\end{equation}
% but in a Siamese architecture \cite{zagoruyko2015learning}
where $r(\cdot)$ is a scoring function (and we implement it with the same structure as $h(\cdot)$).   
We use the binary labels $a_{ij}$ indicating whether two utterances $x_i$ and $x_j$ have the same semantic latent action to train the SPN. 
Specifically, we have $a_{ij}=1$ if $z_d(x_i) = z_d(x_j)$, and otherwise $a_{ij}=0$.
% \begin{equation*}
%           a_{ij}=\begin{cases}
%               1, z_d(x_i) = z_d(x_j)\\
%               0, \text{otherwise}
%             \end{cases}
% \end{equation*}
To facilitate effective knowledge transfer, we obtain such labels from both source and target domains.
We consider all pairs of source domain utterances and obtain
% and the semantic latent actions learned at stage-\RNum{1} and obtain
\begin{equation*}
    P^{\mathcal{S}} = \{ \big \langle (x_i, x_j),a_{ij}  \big \rangle \mid x_i, x_j \in U^{\mathcal{S}}\}.
\end{equation*}
We also consider pairs of target domain utterances with shared actions:
we first get all target domain utterances with aligned actions
$    
U_{\text{shared}}^{\mathcal{T}} = \{x_j^{\mathcal{T}} | x_j^{\mathcal{T}} \in U^{\mathcal{T}}, z_d(x_j^{\mathcal{T}}) \in A^{\mathcal{S}} \}
$ 
where $A^{\mathcal{S}}$ represents the set of shared actions
$
     A^{\mathcal{S}}  = \{z_d(x_i^{\mathcal{S}}) \mid x_i^{\mathcal{S}} \in U^{\mathcal{S}}\} 
$
and then obtain
\begin{equation*}
    P^{\mathcal{T}} = \{\big \langle (x_i, x_j ), a_{ij} \big \rangle  \mid x_i, x_j \in U_{\text{shared}}^{\mathcal{T}} \}.
\end{equation*}
Using all the collected pairwise training instances $p =\big \langle (x_i, x_j ), a_{ij} \big \rangle $, we train SPN via the loss
\begin{equation}
    \mathcal{L}_{\text{SPN}} = \mathbb{E}_{p \in P^{\mathcal{S}} + P^{\mathcal{T}} } [\text{cross-entropy}(a_{ij}, r(x_i,x_j))].
\end{equation}

We then use the trained $p_{\text{SPN}} $ to replace state tracking models in both pointwise and pairwise measure. 
% measure whether two utterances can lead to similar state transitions.
% We can apply this measurement in both pointwise and pairwise manner. 
Specifically, we formulate the following pointwise loss
% The pointwise loss enforces the reconstructed utterances to bring similar dialogue state transitions as the original utterance
% i.e., between reconstructed utterance from VAE and the original one   
\begin{equation}
    \mathcal{L}_{\text{PT}}^{\mathcal{T}} = \mathbb{E}_{x \in U^T}[-\log p_{\text{SPN}}(x^{\mathcal{T}}, \Tilde{x }^{\mathcal{T}}  ) ]
\end{equation}
\begin{equation*}
    \Tilde{x }^{\mathcal{T}} \sim p_{\mathcal{G}} (z_q(x^{\mathcal{T}} ))
\end{equation*}
which enforces the reconstructed utterances to bring similar dialogue state transitions as the original utterance.
We further formulate the pairwise loss as  
\begin{equation}
\begin{split}
   \mathcal{L}_{\text{PR}}^{\mathcal{T-T}} &=  \sum_{x_i,x_j \in U^{\mathcal{T}}} -p_{ \text{SPN}}(x_i,x_j) \log d(x_i^{\mathcal{T}}, x_j^{\mathcal{T}}) \\
   &- (1-p_{ \text{SPN}}(x_i,x_j))\log(1- d(x_i^{\mathcal{T}}, x_j^{\mathcal{T}}))
\end{split}
\end{equation}
\begin{equation*}
      d(x_i^{\mathcal{T}}, x_j^{\mathcal{T}}) = \sigma( -z_e(x_{i}^{\mathcal{T}})^{\top} z_e(x_{j}^{\mathcal{T}} ) ).
\end{equation*}
Compared to the pairwise loss at stage-I (Eqn. \ref{pair-s1}) and stage-II (Eqn. \ref{pair-s2}), the main difference is that we use $p_{\text{SPN}}$ to substitute $s_{i,j}$ that relies on trained dialogue state tracking models.
% we can find that the key idea of stage-III is to use

The overall objective function for stage-\RNum{3} is:
\begin{equation}
    \mathcal{L}_{\text{S-\RNum{3}}} = \mathcal{L}_{\text{a-e}}^{\mathcal{T}} +\alpha \mathcal{L}_{\text{PT}}^{\mathcal{T}} + \beta \mathcal{L}_{\text{PR}}^{\mathcal{T-T}}
\end{equation}

% \begin{equation}
%     \mathcal{L}_{\text{SPN}} = \mathbb{E}_{p \in P } [-a_{ij}\log r(p) -(1-a_{ij})\log (1-r(p) ) ]
% \end{equation}

\subsection{Conditioned Response Generation}
After obtaining semantic latent actions, we train the two components, dialogue planning and surface realization, for conditioned response generation.
Specifically, we first train a surface realization model $p_{r}$ that learns how to translate a semantic latent action into fluent text in context $c$ as 
$$
\mathcal{L} = \mathbb{E}_{x}[- \log p_{r}(x|z_d(x),c  )    ].
$$
% where $z_d(x)$ is the learned semantic latent actions.
%  related to utterance $x$.
% where $c$ is dialogue context. 
Then we optimize a dialogue planning model $p_{l}$ while keeping the parameters of $p_{r} $ fixed 
$$
\mathcal{L} = \mathbb{E}_{x}\mathbb{E}_{z} [- \log p_{r}(x|z,c)p_{l}(z|c) ]
$$
In this way, the response generation is factorized into $p(x|c)=p(x|z,c)p(z|c) $, where dialogue planning and surface realization are optimized without impinging the other.
\section{Experiments}

% Overview paragraph.
To show the effectiveness of MALA, we consider two experiment settings: multi-domain joint training and cross-domain response generation (Sec. 5.1). 
We compare against the state-of-the-art on two multi-domain datasets in both settings (Sec. 5.2).
We analyze the effectiveness of semantic latent actions and the multi-stage strategy of MALA under different supervision proportion (Sec. 5.3).

\subsection{Settings}

% on two multi-domain human-human conversational datasets. 
\subsubsection{Datasets} We use two multi-domain human-human conversational datasets: 
% and has an average of 5.3 turns per dialogue.
(1) \textsc{SMD} dataset \cite{eric2017key} contains 2425 dialogues, and has three domains: \textit{calendar}, \emph{weather}, \emph{navigation}; 
(2) \textsc{MultiWOZ} dataset \cite{budzianowski2018multiwoz} is the largest existing task-oriented corpus spanning over seven domains. It contains in total 8438 dialogues and each dialogue has 13.7 turns in average.
We only use five out of seven domains, i.e., \emph{restaurant, hotel, attraction, taxi, train}, since the other two domains contain much less dialogues in training set and do not appear in testing set.
This setting is also adopted in the study of dialogue state tracking transferring tasks \cite{wu2019transferable}.
% Both datasets are created by a Wizard-of-Oz method, means that.
Both datasets contain dialogue states annotations.
% One key difference between these two datasets is that \textsc{MultiWOZ} involves a much larger dialogue action space, and is thus much more challenging. 

% Evaluation

% We use automatic evaluation metrics to evaluate dialogue task completion. 
We use \textbf{Entity-F1} \cite{eric2017key} to evaluate dialogue task completion, which computes the F1 score based on comparing entities in delexicalized forms.
Compared to inform and success rate originally used on \textsc{MultiWOZ} by \mbox{Budzianowski et al. (2018)}, Entity-F1 considers informed and requested entities at the same time and balances the recall and precision.   
% We also include the results under inform rate and success rate in Appendix.
We use \textbf{BLEU} \cite{papineni2002bleu} to measure the language quality of generated responses.
We use a three-layer transformer \cite{vaswani2017attention} with a hidden size of 128 and 4 heads as base model.
\iffalse
On \textsc{MultiWOZ}, we use Inform Rate and Success Rate as in the Dialog-Context-to-Text Generation task proposed by Budzianowski et al. (2018): \textbf{Inform} rate measures whether the system has provided an appropriate entity; \textbf{Success} rate measures whether generated responses answer all the requested attributes.
\fi

\begin{table} [tbp]
    \centering

\begin{threeparttable}
\caption{ Multi-Domain Joint Training Results 
} \label{multi-join}

% \tiny
%\scriptsize
%\footnotesize
\small

\setlength\tabcolsep{3.0pt}

\begin{tabular}{l|l|cc||cc}
\toprule
 \multicolumn{2}{c|}{} & \multicolumn{2}{c||}{ \textsc{SMD}} & \multicolumn{2}{c}{ \textsc{MultiWOZ     } } \\
\cmidrule{3-4}\cmidrule{5-6}
\multicolumn{2}{c|}{\textsc{Model} }& Entity-F1 & BLEU    & Entity-F1 & BLEU   \\
\midrule
\multirow{3}{*}{ \makecell{w/o Action} } &KVRN   & 48.1  & 13.2  & 30.3  & 11.3  \\
& Mem2seq   & 62.6 & 20.5   & 39.2  & 14.8  \\
& Sequicity   & 81.1 & 21.9 & 57.7   & 17.2   \\
\midrule
\multirow{2}{*}{ \makecell{w/ Action} } & LIDM   &	76.7 & 17.3 & 59.4  & 15.5   \\
& LaRL   &	80.4 & 18.2  & 71.3 & 14.8   \\
\midrule
\multirow{3}{*}{ \makecell{Proposed} }&MALA-S1    & 83.8 & 22.4 & 74.3    & 18.7  \\
& MALA-S2   & 84.7 & 21.7 & 76.2   & 20.0  \\
& MALA-S3    &	\textbf{85.2} & \textbf{22.7}   & \textbf{76.8} & \textbf{20.1}   \\
\bottomrule
\end{tabular}
\begin{tablenotes}\footnotesize
\item[*] Note that w/o and w/ Action means whether the baseline considers conditioned generation
\end{tablenotes}
\end{threeparttable}

\end{table}

\begin{table} [tbp]
\begin{threeparttable}

\caption{ Cross-Domain Generation Results on \textsc{SMD} 
} \label{cross-smd}
    \centering

% \tiny
%\scriptsize
% \footnotesize
\small

\setlength\tabcolsep{3.5pt}

\begin{tabular}{l|l|c|c|c|c}
\toprule
% \multicolumn{2}{c|}{} & \multicolumn{2}{c|}{Navigation} & \multicolumn{2}{c}{ Weather }  \\
%\multicolumn{2}{c|}{ \multirow{2}{*}{ \textsc{Model}}  }  & \multirow{2}{*}{BLEU}  &  \multicolumn{3}{c}{Enty-F1 on target domain} \\
\multicolumn{2}{c|}{} & \multicolumn{3}{c|}{Entity-F1 in target domain} & \multirow{2}{*}{\makecell{\\BLEU}}\\
\cmidrule{3-5}
\multicolumn{2}{c|}{\textsc{Model}}  & Navigate & Weather & Schedule &  \\
\midrule
\multirow{2}{*}{ \makecell{Target \\ Only} } & Sequicity   & 31.7   & 42.6 & 55.7 & 16.0 \\
& LaRL   & 33.2   & 44.3 & 57.5 & 12.3 \\
\midrule
\multirow{2}{*}{ \makecell{Fine \\ Tuning} }& Sequicity   &	35.9 & 46.9 & 59.7 & 16.8\\
& LaRL   &	34.7 & 45.0 & 58.6 & 12.1\\
\midrule
\multirow{3}{*}{ Proposed} & MALA-S1    &	38.3 & 54.8 & 64.4 & 19.3 \\
& MALA-S2    &	39.4 & 57.0 & 65.1 & 18.5  \\
& MALA-S3    &	\textbf{41.8} & \textbf{59.4} & \textbf{68.1} & \textbf{20.2} \\
\bottomrule
\end{tabular}

\end{threeparttable}

\end{table}

\begin{table*} [tbp]
    \centering

\begin{threeparttable}

\caption{ Cross-Domain Generation Results on \textsc{MultiWOZ} 
} \label{cross-woz}

% \tiny
% \scriptsize
% \footnotesize
\small

\setlength\tabcolsep{3.5pt}

\begin{tabular}{l|l|cc|cc|cc|cc|cc}
\toprule
% Method & nDCG & $\alpha\mbox{-}$nDCG & p$\mbox{-}$nDCG & nDCG & $\alpha\mbox{-}$nDCG & p$\mbox{-}$nDCG \\
\multicolumn{2}{c|}{} & \multicolumn{2}{c|}{Hotel} & \multicolumn{2}{c|}{ Train } &  \multicolumn{2}{c|}{ Attraction }& \multicolumn{2}{c|}{Restaurant}& \multicolumn{2}{c}{Taxi} \\
\cmidrule{3-12}
\multicolumn{2}{c|}{ \textsc{Model}} & Entity-F1  & BLEU &  Entity-F1  & BLEU &  Entity-F1  & BLEU &  Entity-F1  & BLEU &  Entity-F1  & BLEU\\
\midrule
\multirow{2}{*}{ Target Only } & Sequicity & 16.1  & 10.7   & 27.6 &16.8  &  17.4 & 14.4 & 19.6& 13.9 & 22.1 & 15.4\\
& LaRL & 17.8  & 10.1 & 30.5 & 12.9  &  24.2 & 11.7  & 19.9 & 9.6  & 28.5 & 11.7 \\
\midrule
\multirow{2}{*}{ Fine Tuning }& Sequicity  & 17.3  &12.3	 &27.0  &17.6  &17.9  &15.8   &26.0  &14.5   &22.4  &16.9  \\
& LaRL & 21.0  &9.1	 &34.7  &12.8  &24.8  &11.8   &22.1  &10.8   &31.9  &12.6  \\
\midrule
\multirow{3}{*}{ Proposed} & MALA-S1   &23.3   &15.5	 &43.5  &18.1  &31.5  &16.2   &24.7  &16.5   &33.6  &18.0 \\
& MALA-S2  & 26.4  &	15.8 & 48.3 & 18.8 &36.5  &17.6   &28.8  &16.6   &41.7  &18.6  \\
& MALA-S3  & \textbf{32.7}  &\textbf{16.7}	 & \textbf{51.2} & \textbf{19.4} & \textbf{41.9} &\textbf{18.1}   &\textbf{35.0}  &\textbf{17.3}   &\textbf{44.7}  &\textbf{19.0}\\
\bottomrule
\end{tabular}

\end{threeparttable}

\end{table*}

\subsubsection{Multi-domain Joint Training}
In this setting, we train MALA and other baselines with full training set, i.e., using complete dialogue data and dialogue state annotations.
We use the separation of training, validation and testing data as original \textsc{SMD} and \textsc{MultiWOZ} dataset. We compare with the following baselines that do not consider conditioned generation: 
(1) \textbf{KVRN} \cite{eric2017key};
(2) \textbf{Mem2seq} \cite{madotto2018mem2seq};
(3) \textbf{Sequicity} \cite{lei2018sequicity};
% We also consider two baselines that adopts latent action learning for conditioned response generation:
and two baselines that adopt conditioned generation:
(4) \textbf{LIDM} \cite{wen2017latent};
(5) \textbf{LaRL} \cite{zhao2019rethinking};
For a thorough comparison, We include the results of the proposed model after one, two, and all three stages, denoted as \textbf{MALA-(S1/S2/S3)}, in both settings.

\subsubsection{Cross-domain Response Generation}
In this setting, we adopt a leave-one-out approach on each dataset. 
Specifically we use one domain as target domain while the others as source domains.
There are three and five possible configurations for \textsc{SMD} and \textsc{MultiWOZ}, respectively.
For each configuration, we set that only 1\% of dialogues in target domain are available for training, and these dialogues have no state annotations. 
We compare with Sequicity and LaRL using two types of training schemes in cross-domain response generation.~\footnote{We also consider using DAML \cite{qian2019domain}, but the empirical results are worse than those of target only and fine tuning.} 
(1) Target only: models are trained only using dialogues in target domain.
(2) Fine tuning: model are first trained in the source domains, and we conduct fine-tuning using dialogues in target domain.

% For a thorough comparison, We include the experiment results of the proposed model after one, two, and all three stages, denoted as MALA-(S1/S2/S3), in both settings.

\subsection{Overall Results}
% We compare MALA with multiple baselines in both task completion and language generation quality.
\subsubsection{Multi-Domain Joint Training}
Table \ref{multi-join} shows that our proposed model consistently outperforms other models in the joint training setting.
MALA improves dialogue task completion (measured by Entity-F1) while maintaining a high quality of language generation (measured by BLEU).
For example, MALA-S3 (76.8) outperforms LaRL (71.3) by 7.71\% under Entity-F1 on \textsc{MultiWOZ}, and has the highest BLEU score. 
% Among our proposed models, MALA-S3 achieves the best performance on both datasets, meaning that the multi-stage strategy  in the joint-training setting.
Meanwhile, we also find that MALA benefits much from \mbox{stage-I} and \mbox{stage-II} in the joint learning setting.
For example, MALA-S1 and MALA-S2 achieve 9.25\% and 10.43\% improvements over LIDM under Entity-F1 on \textsc{SMD}.
This is largely because that, having complete dialogue state annotations, MALA can learn semantic latent actions in each domain at stage-I, and the action alignment at stage-II reduces the action space and thus further enhances the effectiveness of dialogue policy learning.
% the learned similarity measure at the first two stages can well learn domain-specific actions.
% In general, models that consider conditioned generation outperform the others.
We further find that LIDM and LaRL perform worse than Sequicity on \textsc{SMD}.
The reason is that system utterances on \textsc{SMD} have shorter lengths and various expressions, making it challenging to capture underlying intentions merely based on surface realization.
MALA overcomes this challenge by considering dialogue state transitions beyond surface realization in semantic latent action learning.
% with semantic latent action learning 
% and achieve the best Entity-F1
% by encoding dialogue state transitions caused  
% by system utterances in semantic latent actions.
% thus achieve the best Entity-F1 and BLEU on \textsc{SMD}.

% With semantic latent action learning, 

% MALA successfully overcomes such challenge and performs the best on \textsc{SMD}.

\subsubsection{Cross-Domain Response Generation} 
The results on \textsc{SMD} and \textsc{MultiWOZ} are shown on Tables \ref{cross-smd} and \ref{cross-woz}, respectively.
We can see that MALA significantly outperforms the baselines on both datasets.
For example, on \mbox{\textsc{MultiWOZ}}, MALA-S3 outperforms LaRL by 47.5\% and 55.7\% under Entity-F1 using \textit{train} and \textit{hotel} as target domain, respectively.
% the proposed multi-stage strategy suits the cross-domain generation scenarios, 
We also find that each stage of MALA is essential in the cross-domain generation scenario.
For example, on \textsc{MultiWOZ} using \textit{attraction} as target domain, stage-III and stage-II brings 14.7\% and 15.8\% improvements compared with its former stage, and MALA-S1 outperforms fine-tuned LaRL by 27.0\% under Entity-F1.
% using \textit{attraction} as target domain.
% detail in the following subsections
We further find that the contribution of each stage may vary when using different domains as target, and we will conduct a detailed discussion in the following section.
By comparing fine-tuning and target only results of LaRL, we can see latent actions based on lexical similarity cannot well generalize in the cross-domain setting.  
For example, fine-tuned LaRL only achieves less than 3\% improvement over target-only result under Entity-F1 on MultiWOZ using \textit{attraction} as target domain.

\subsection{Discussions}
\begin{figure}[t]
\centering
\subfigure[\small{Restaurant as target domain}]{
\begin{overpic}[height=3.56cm]{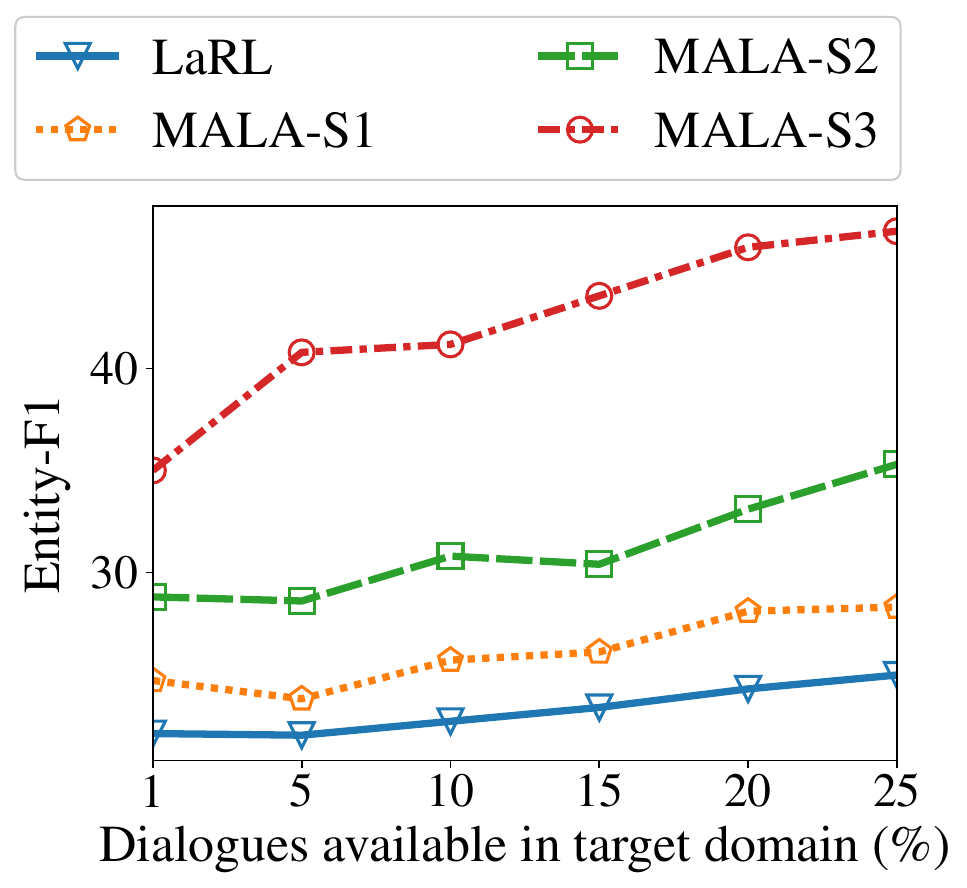}
\label{target-restau}
\end{overpic}
}
\subfigure[\small{Taxi as target domain}]{
\begin{overpic}[height=3.56cm]{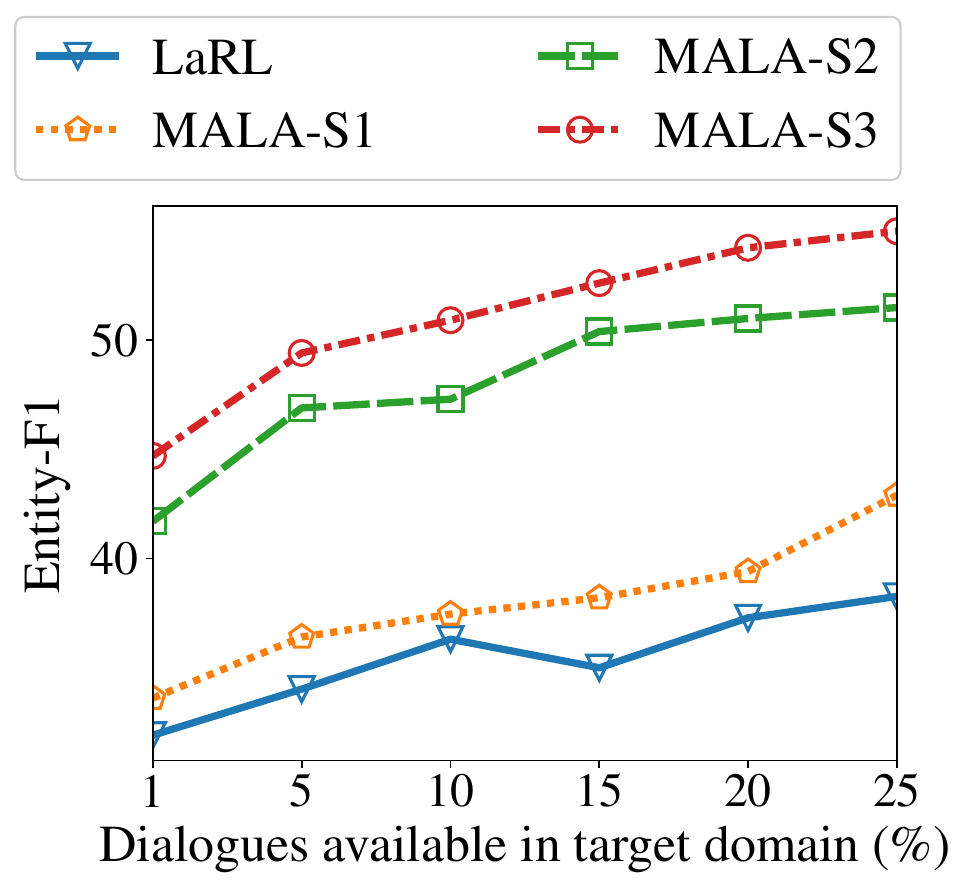}
\label{target-taxi}
\end{overpic}
}
\caption{Effects of multiple stages on \textsc{MultiWOZ}}
\label{effect-stage}
\end{figure}

% \subsubsection{Effects of Multi-Stage Strategy}
We first study the effects of each stage in MALA in cross-domain dialogue generation.
We compare MALA-(S1/S2/S3) with fine-tuned LaRL under different dialogue proportions in the target domain.
% of all models under Entity-F1 on MultiWOZ 
The results are shown in Fig. \ref{target-restau} and \ref{target-taxi}. 
% As expected, all models perform better with the increase of available dialogues on target.  
We can see that the performance gain of MALA is largely attributed to stage-III when using \textit{restaurant} as target domain, while attributed to stage-II using \textit{taxi} as target.
% and when using \textit{taxi} as target domain, 
This is largely because there are many shared actions between the \textit{taxi} and \textit{train} domain, and thus many utterance-action pairs learned by action alignment at stage-II already capture the underlying intentions of utterances.
% On the other hand, when using \textit{restaurant} as target domain, since there are not many shared actions found across domains, MALA relies much on the utterance-level similarity prediction network at stage-III.
On the other hand, since \textit{restaurant} does not have many shared actions across domains, MALA relies more on the similarity prediction network to provide supervision at stage-III.

\begin{figure}[!t]
\centering
\subfigure[\small{Multi-domain joint training}]{
\begin{overpic}[height=3.56cm]{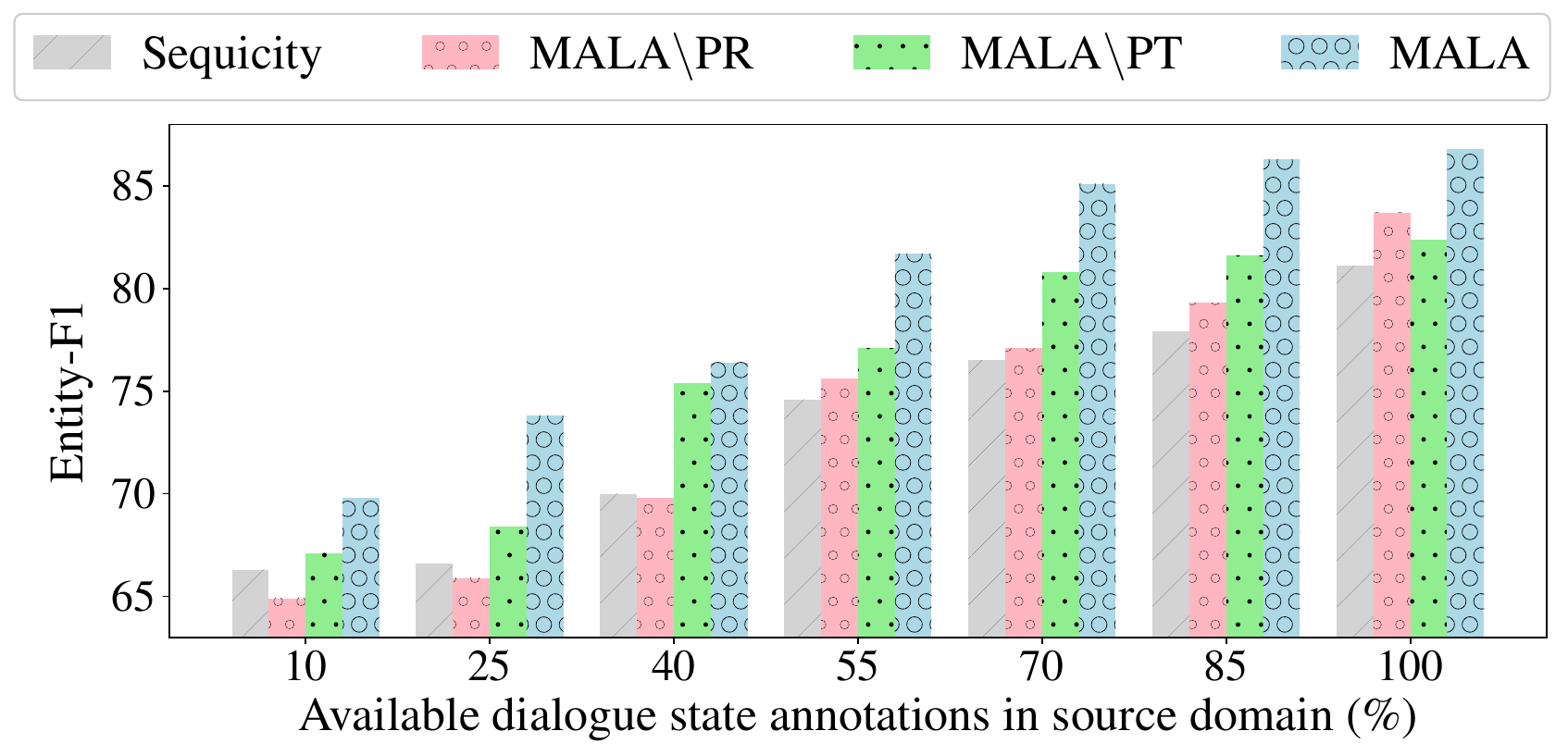}
\label{state-mul}
\end{overpic}
}

\subfigure[\small{Cross-domain generation, navigation as target domain}]{
\begin{overpic}[height=3.86cm]{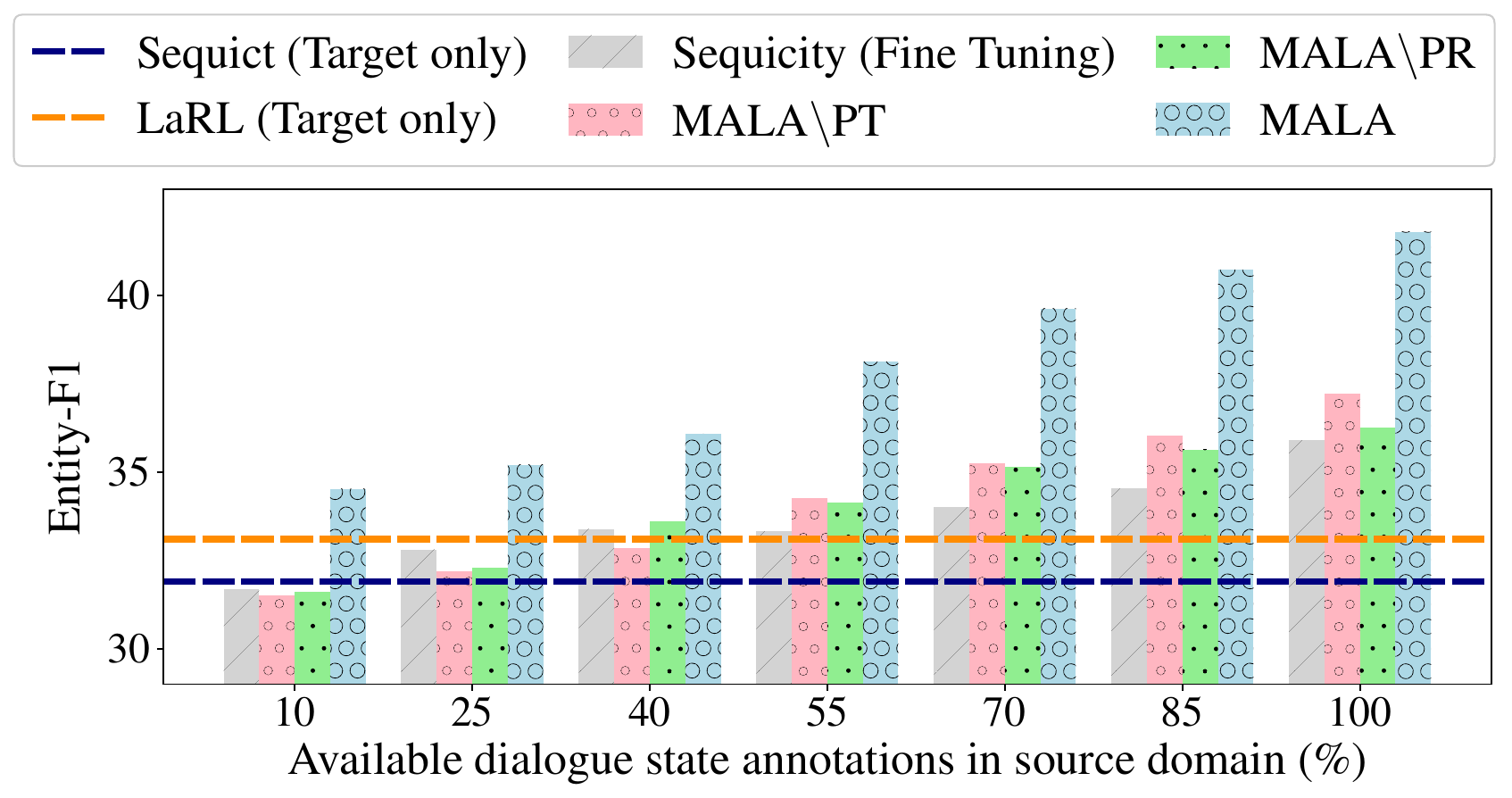}
\label{state-target}
\end{overpic}
}
\caption{Effects of semantic action learning on \textsc{SMD}}
\label{effect-semantic}
\end{figure}

% while \textit{restaurant} as target doesn't hold such property.   
% These results show that MALA can well learn actions on target domains whether there are many shared actions.

% \subsubsection{Effects of Semantic Action Learning}
Lastly, we study the effects of semantic latent actions in both the joint training and the cross-domain generation setting.
% At Stage-I of MALA, we propose to use pointwise effect measure together pairwise similarity measure to learn semantic actions.
% Recall that we encode pointwise measure $\mathcal{L}_{\text{PT}}$ and pairwise measure $\mathcal{L}_{\text{PR}}$ into semantic latent action learning.   
To investigate how pointwise measure $\mathcal{L}_{\text{PT}}$ and pairwise measure $\mathcal{L}_{\text{PR}}$ contribute to capturing utterance intentions, we compare the results of MALA without pointwise loss (MALA$\setminus$PT), and without pairwise loss (MALA$\setminus$PR) under varying sizes of dialogue state annotations.
The results of multi-domain joint training under Entity-F1 on SMD are shown in Fig. \ref{state-mul}.
We can see that both pointwise and pairwise measure are important.
For example, when using 55\% of state annotations, encoding pointwise and pairwise measure bring 5.9\% and 8.0\% improvement, respectively.
For cross-domain generation results shown in Fig. \ref{state-target}, we can find that these two measures are essential to obtain semantic latent actions in the target domain.
% we can find that learning semantic latent action is also essential for cross-domain scenarios.  
% For example, pointwise and pairwise measure bring improvements when having full state annotations.

\section{Conclusion}
We propose multi-stage adaptive latent action learning (MALA) for better conditioned response generation.
% for cross-domain dialogue generation.
We develop a novel dialogue state transition measurement for learning semantic latent actions.   
We demonstrate how to effectively generalize semantic latent actions to the domains having no state annotations.    
The experimental results confirm that MALA achieves better task completion and language quality compared with the state-of-the-art under both in-domain and cross-domain settings.
For future work, we will explore the potential of semantic action learning for zero-state annotations application.

\clearpage
\section*{Acknowledgement}
We would like to thank Xiaojie Wang for his help.
This work is supported by Australian Research Council (ARC) Discovery Project DP180102050.
{
\bibliographystyle{aaai}
\bibliography{aaai20.bib}
}

\end{document}